%% file: main.tex
\documentclass{article}

\usepackage[preprint]{neurips_2026}

\usepackage[utf8]{inputenc} 
\usepackage[T1]{fontenc}    
\usepackage{microtype}
\usepackage{graphicx}
\usepackage{booktabs}       
\usepackage{multirow}       
\usepackage[table]{xcolor}  
\usepackage{caption}        
\usepackage{hyperref}
\usepackage{url}
\usepackage{amsfonts}       
\usepackage{nicefrac}       

\usepackage{amsmath}
\usepackage{amssymb}
\usepackage{mathtools}
\usepackage{amsthm}

\usepackage[capitalize,noabbrev]{cleveref}

\usepackage[textsize=tiny]{todonotes}

\theoremstyle{plain}

\theoremstyle{definition}

\theoremstyle{remark}

\title{Rethinking Dense Sequential Chains: Reasoning Language Models Can Extract Answers from Sparse, Order-Shuffling Chain-of-Thoughts}

\author{%
  Yi-Chang Chen$^{1}$\thanks{Correspondence: \texttt{ycc.tw.email@gmail.com}} \quad
  Feng-Ting Liao$^{1}$ \quad
  Da-shan Shiu$^{1}$ \quad
  Hung-yi Lee$^{2}$ \\[0.3em]
  {\small
    $^{1}$MediaTek Research \qquad
    $^{2}$Artificial Intelligence Center of Research Excellence, National Taiwan University
  }
}

\begin{document}

\maketitle

\input{sections/abstract}

\input{sections/introduction}

\input{sections/related_work}

\input{sections/methodology}

\input{sections/results}

\input{sections/conclusion}


\bibliography{references}
\bibliographystyle{plainnat}

\newpage
\appendix

\input{tables/remove_alphabet_examples}

\end{document}

%% file: sections/abstract.tex

\begin{abstract}
Modern reasoning language models generate dense, sequential chain-of-thought traces implicitly assuming that every token contributes and that steps must be consumed in order. We challenge both assumptions through a systematic intervention pipeline—removal, masking, shuffling, and noise injection—applied to model-generated reasoning chains across three models of different sizes and from different model families, evaluated on three challenging benchmarks spanning distinct domains.
Our findings are counterintuitive on three dimensions.
\emph{Order}: Does the sequential order of a reasoning chain matter for answer extraction?
No---line-level shuffling reduces accuracy by less than 0.5 percentage points (pp); word-level shuffling retains 62\%--89\% accuracy; only token-level shuffling collapses to near zero.
Crucially, pretrained-only and instruction-tuned variants of the same model exhibit near-identical tolerance to both perturbations (78.67\% vs.\ 78.00\% under line shuffling), indicating this order-independence originates from pretraining rather than reasoning-specific fine-tuning.
\emph{Dense}: Is all the information in a reasoning chain important for answer extraction?
No---In our mathematical reasoning experiments, only symbolic and numeric content (including the explicit answer occurrence) is irreducible.
Masking numeric digits collapses accuracy to exactly 0\%, while masking alphabetic prose even improves accuracy by 4.7 pp.
\emph{Robustness}: Is a reasoning chain that is both order-shuffling and non-dense still robust?
Yes---the most aggressively reduced representation (all natural language removed, lines arbitrarily shuffled) still achieves 83\% accuracy, and injecting false answers at $3\times$ true-answer frequency leaves accuracy entirely unchanged (83.3\%\,$\to$\,83.3\%), falsifying a frequency-based extraction account.
Together, these results establish that answer extraction operates on a sparse, order-shuffling, and structurally robust informational substrate, opening paths toward parallelized and token-efficient reasoning generation.
All code and data are publicly available.\footnote{\url{https://github.com/mtkresearch/reasoning-behavior}}
\end{abstract}

%% file: sections/introduction.tex

\section{Introduction}
\label{sec:introduction}

Large reasoning language models---such as OpenAI's o1 \citep{openai2024o1} and DeepSeek-R1 \citep{deepseek2024r1}---achieve strong performance on challenging multi-step tasks by generating extended chain-of-thought (CoT) traces \citep{wei2022chain} before producing a final answer.
At inference time, the answer-generation model conditions on this entire trace to extract the final response.

This paradigm rests on two implicit assumptions.
The first is \emph{density}: every token in the reasoning trace contributes meaningfully to the final answer.
The second is \emph{order}: reasoning steps must be produced and consumed sequentially.
Together these assumptions make reasoning inherently expensive: inference cost grows linearly with problem difficulty, and auto-regressive dependency prevents parallelization.

Neither assumption has been empirically validated for the answer-\emph{extraction} stage.
If extraction depends only on a sparse, unordered subset of the trace's content, then the extraction stage imposes neither an ordering constraint nor a completeness constraint on what the generation stage produces---and the two stages could in principle be decoupled.
Such decoupling would liberate generation from its current constraints: if the extraction stage can tolerate order-shuffling reasoning chains, reasoning steps need not be produced auto-regressively and could instead be generated in parallel; if the extraction stage can tolerate sparse reasoning chains, intermediate tokens need not be produced exhaustively and could instead be selectively omitted.
We therefore ask: \emph{what is the actual information structure a model relies upon when extracting an answer from a reasoning trace?}

To answer this empirically, we apply controlled transformations---\textbf{removal} (alphabetic characters, question text), \textbf{masking} (alphabetic text, numeric digits, answer occurrences), \textbf{shuffling} (tokens, words, lines, and within-line words), and \textbf{noise injection}---to model-generated reasoning chains and measure the resulting change in answer accuracy across three models from different organizations and parameter scales (\texttt{GPT-OSS-120B}, \texttt{DeepSeek-V3.1-671B}, \texttt{OLMo-3.1-32B}) and three benchmarks covering three distinct reasoning tasks---mathematical computation (\texttt{AIME 2025}), algorithmic problem solving (\texttt{CodeElo}), and scientific inference (\texttt{GPQA-Diamond})---verifying that our observations are not attributable to a single model's characteristics or a single domain.

Our findings are counterintuitive.
\textbf{On order-shuffling}: randomly permuting the \emph{lines} of a reasoning chain reduces accuracy by less than 0.5 percentage points (pp) across all models and all three domains; word-level shuffling retains 62\%--89\% accuracy on \texttt{AIME 2025} while token-level shuffling collapses to near zero, establishing word-level semantic identity as the minimum necessary granularity for extraction. Crucially, we found that pretrained-only and instruction-tuned variants of the same model exhibit near-identical tolerance to both line-level and word-level shuffling, indicating this property is rooted in the transformer's pretraining regime rather than reasoning-specific fine-tuning.
\textbf{On sparsity}: removing all alphabetic characters from a mathematical reasoning chain costs only 1.3 pp of accuracy; masking alphabetic text actually \emph{improves} accuracy by 4.7 pp, indicating natural-language prose is not the key element for answer extraction in this setting; a chain stripped of all natural language and shuffled into arbitrary line order still achieves 83\% accuracy.
\textbf{On robustness}: injecting false-answer sentences at three times the true-answer frequency leaves accuracy entirely unchanged (91.3\% $\to$ 91.3\%).
These results establish that answer extraction is \emph{sparse}, \emph{order-shuffling}, and \emph{structurally robust}---directly contradicting both assumptions of the current paradigm.

Taken together, these findings suggest that transformers process reasoning chains more like \emph{unordered sets of symbolic constraints} than sequential proofs---attending selectively to numeric relational structures while largely ignoring prose and positional arrangement.
This has direct practical implications: if extraction tolerates both reordering and content sparsity, the generation stage could in principle be freed from its current auto-regressive, exhaustive constraints---reasoning steps could be generated in parallel, and uninformative prose tokens could be selectively omitted.

This work makes the following contributions:
\begin{itemize}
    \item \textbf{A systematic empirical framework for probing answer extraction.} A suite of controlled transformations (removal, masking, shuffling, noise injection) across three models (\texttt{GPT-OSS-120B}, \texttt{DeepSeek-V3.1-671B}, \texttt{OLMo-3.1-32B}) and three domains (\texttt{AIME 2025}, \texttt{CodeElo}, \texttt{GPQA-Diamond}), enabling reproducible, fine-grained isolation of which information the extraction stage actually uses.
    \item \textbf{Reasoning chains are order-shuffling at the line level; word-level identity is the minimum necessary granularity.} Line-order destruction causes negligible accuracy loss ($<$0.5 pp); word-level semantic identity is the minimum necessary unit for extraction---suggesting reasoning steps need not be presented, or perhaps generated, sequentially.
    \item \textbf{Order-independence is present in pretrained models.} Pretrained and instruction-tuned variants show near-identical shuffling tolerance, ruling out reasoning-specific fine-tuning as the source of this property.
    \item \textbf{Informational content is non-uniformly distributed---alphabetic prose is redundant while numeric content and the explicit answer are irreducible.} A chain stripped of all natural language and presented in arbitrary line order still achieves 83\% accuracy.
    \item \textbf{The structural signal is intrinsically robust---even strong false-answer injection cannot override it.} Extraction is unaffected at $3\times$ true-answer frequency, indicating the signal is governed by relational constraint satisfaction over numeric structures rather than surface-level frequency counting.
\end{itemize}

%% file: sections/related_work.tex

\section{Related Work}
\label{sec:related}

A body of prior work has examined what information in reasoning traces actually drives model performance.
\citet{min2022rethinking} show that the correctness of in-context demonstration labels contributes surprisingly little to task performance---the label space, input distribution, and sequence format matter far more than literal semantic content.
\citet{madaan2022text} decompose CoT prompts into structural patterns and semantic content, finding that patterns drive performance while factual content is largely expendable; their Concise CoT achieves over 20\% token reduction with minimal accuracy loss.
\citet{lanham2023measuring} show through intervention experiments that model reliance on the reasoning trace is highly task-dependent, and that larger models tend to produce \emph{less} faithful CoT.
\citet{pfau2024think} demonstrate that transformers can achieve strong performance on demanding tasks even when intermediate tokens are meaningless fillers---the primary mechanism is the extra computational budget, not semantic content.
Together, these results suggest that CoT reasoning chains contain substantial redundancy, and that computation space rather than semantic content may be the primary driver of performance.

Our work extends this line by directly intervening on model-generated reasoning chains and quantifying information geometry along two orthogonal axes: sparsity and order-dependence.
Unlike prior work that modifies human-authored demonstrations or analyzes model behavior indirectly, we apply a configurable transformation pipeline to chains produced by frontier models and measure the resulting accuracy change across multiple models and domains.
Removing all natural-language text from model-generated reasoning chains costs less than 2 pp of accuracy, and reordering lines causes negligible degradation, suggesting that sequential structure is less critical than the presence of the right informational tokens anywhere in the context.

%% file: sections/methodology.tex

\section{Methodology}
\label{sec:methodology}

\subsection{Problem Formulation}
\label{sec:formulation}

A traditional reasoning model generates an answer $a$ given a question $q$ and
reasoning chain $r = (r_1, r_2, \ldots, r_n)$:
\begin{equation}
  P(a \mid q, r)
\end{equation}
Our central hypothesis is that answer generation can succeed with a
\emph{sparse and order-shuffling} representation $\tilde{r} = T(r;\,\theta)$:
\begin{equation}
  P(a \mid q, r) \approx P(a \mid q, \tilde{r})
\end{equation}
where $T$ controls the degree of sparsity or reordering applied to $r$.
We operationalize this hypothesis by measuring answer accuracy under a suite
of controlled transformations applied to model-generated reasoning chains,
using an untransformed chain as the baseline.

\input{tables/transformation_processors}

\subsection{Experimental Setup}
\label{sec:setup}

Our experimental pipeline proceeds in three stages.

\textbf{Stage 1 - Reasoning chain collection.}
Before any intervention, we run each model (\Cref{sec:models}) on each
benchmark (\Cref{sec:datasets}) under its standard reasoning mode to collect
model-generated reasoning chains; these untransformed chains serve as the
substrate for all subsequent transformations.

\textbf{Stage 2 - Transformation and measurement.}
To probe the hypothesis in \Cref{sec:formulation}, we apply the processors
listed in \Cref{tab:processors} to each collected chain and measure answer
accuracy $\text{Accuracy} = \text{\#correct}/\text{\#success}$.
Processors compose sequentially---e.g.,
$\mathcal{S}_\text{line} \circ \mathcal{R}_{\alpha}$---enabling systematic
exploration of joint transformation effects.
The correctness criterion for each benchmark is described in \Cref{sec:datasets}.

\textbf{Stage 3 - Evaluation.}
We evaluate under two modes.
In \textbf{Free Generation (\textsf{Gen})} mode, the model produces an answer without
any structural constraint beyond the provided (transformed) reasoning chain.
In \textbf{Retrieval (\textsf{Ret})} mode, a task-specific completion prefix is
appended---\textit{``Thus, the answer is''} for \texttt{AIME 2025} and
\texttt{GPQA-Diamond}; \textit{``Thus, the code
is\textbackslash{}n\textasciigrave\textasciigrave\textasciigrave{}cpp\textbackslash{}n''}
for \texttt{CodeElo}---constraining the model to produce an immediate answer
without further deliberation.
We adopt \textbf{\textsf{Ret} as the primary evaluation setting}; the empirical
motivation for this choice is developed in \Cref{sec:results-genret}, with \textsf{Gen}
results included for reference.

\textbf{Baseline comparison.}
The \textbf{baseline} applies no transformation: the original reasoning chain
is provided verbatim, and each transformed result is compared against it.
Baseline accuracies are reported in the \textsf{Original} + \textsf{Ret} row of
\Cref{tab:shuffle_comparison}.

\textbf{Compute resources.}
This work involves no model training; all experiments consist solely of model
inference.
All inference calls are routed through the OpenRouter API\footnote{\url{https://openrouter.ai/}},
which provides unified access to the hosted models used in our study.

\subsection{Datasets}
\label{sec:datasets}

We evaluate primarily on \texttt{AIME 2025} \citep{aime25}, comprising 30 high-difficulty competition problems spanning
algebra, geometry, combinatorics, and number theory.
All answers are integers in the range 0--999.
We sample 10 independent reasoning traces per problem (300 total instances)
to reduce variance.
Correctness is judged by a \texttt{GPT-OSS-120B} judge that checks numerical
equivalence between the extracted answer and the ground truth.

\texttt{CodeElo} \citep{quan2025codeelo} is a competitive programming benchmark
drawn from Codeforces, covering problems across a wide range of difficulty
ratings.
Each problem requires producing executable code that satisfies a formal
specification; correctness is determined by execution against a fixed test suite,
with no judge model required.

\texttt{GPQA-Diamond} \citep{rein2023gpqa} is a graduate-level science
benchmark comprising expert-curated multiple-choice questions (four options:
A/B/C/D) spanning biology, chemistry, and physics.
Questions are designed to resist lookup strategies and require multi-step
scientific reasoning; even domain experts achieve only around 65\% accuracy,
making it a stringent test of deep reasoning.
Correctness is judged by a \texttt{GPT-OSS-120B} judge that identifies the
selected option from the model's response.

The three benchmarks cover three qualitatively distinct reasoning
modalities---mathematical computation, algorithmic problem solving, and
scientific inference---enabling us to assess the generalizability of our
findings across domains.

\subsection{Models}
\label{sec:models}

We use three models to collect reasoning chains (\Cref{sec:setup}):
\texttt{OLMo-3.1-32B} \citep{olmo2025olmo}, \texttt{GPT-OSS-120B} \citep{agarwal2025gpt}, and
\texttt{DeepSeek-V3.1-671B} \citep{deepseek2024v3}.
Each model serves in two roles: as the \textbf{reasoning generator} (producing
chains via its built-in reasoning mode) and as the \textbf{answer extractor}
(consuming the transformed chain to produce a final answer).
This paired design---each model generating and extracting from its own
chains---isolates the effect of the transformation itself rather than any
generator--extractor style mismatch.
We additionally compare \texttt{OLMo-3.1-32B-Base} (pretrained only, without
instruction tuning) against \texttt{OLMo-3.1-32B} to probe whether
order-independence originates from reasoning-specific fine-tuning or from
pretraining.
All answer generation uses temperature 0.5 and a maximum of 5{,}000 output
tokens.

%% file: tables/transformation_processors.tex
\begin{table*}[t]
\caption{Transformation processors used in this work.
         We implement five categories:
         \textbf{Shuffling} ($\mathcal{S}$) permutes content at multiple granularities;
         \textbf{Masking} ($\mathcal{M}$) replaces targeted character classes with a mask token;
         \textbf{Removal} ($\mathcal{R}$) strips content entirely;
         \textbf{Noise Injection} ($\mathcal{N}_k$) inserts false-answer sentences at $k\times$ the true-answer frequency;
         \textbf{Randomisation baselines} ($\mathcal{D}$) replace tokens or words with random samples.
         Processors compose sequentially (e.g., $\mathcal{S}_\text{line} \circ \mathcal{R}_{\alpha}$).}
\label{tab:processors}
\begin{center}
\begin{footnotesize}
\setlength{\tabcolsep}{4pt}
\renewcommand{\arraystretch}{1.0}
\begin{tabular}{@{}rlp{9.8cm}@{}}
\toprule
Symbol & Name & Operation \\
\midrule
$\mathcal{S}_\text{tok}$ & \textsf{Token-Shuffle}     & Permute every subword token globally. \\
$\mathcal{S}_\text{word}$ & \textsf{Word-Shuffle}      & Treat the entire chain as a flat bag of words and permute globally. \\
$\mathcal{S}_\text{line}$  & \textsf{Line-Shuffle}      & Permute newline-delimited segments globally; within-line content is preserved. \\
$\mathcal{S}_\text{ilw}$  & \textsf{In-line-Word-Shuffle} & Independently permute words within each line; line order is preserved. \\
\midrule
$\mathcal{M}_{\alpha}$ & \textsf{Mask-Alphabet} & Replace all alphabetic characters with $\blacksquare$. \\
$\mathcal{M}_{\nu}$    & \textsf{Mask-Number}   & Replace all numeric digits with $\blacksquare$. \\
$\mathcal{M}_\text{ans}$      & \textsf{Mask-Answer}   & Replace all occurrences of the ground-truth answer with $\blacksquare$. \\
\midrule
$\mathcal{R}_{r}$      & \textsf{Reasoning-Free}    & Omit the reasoning chain entirely from the prompt. \\
$\mathcal{R}_{\alpha}$ & \textsf{Remove-Alphabet} & Strip all alphabetic characters; retain digits, symbols, and whitespace. \\
$\mathcal{R}_\text{ans}$      & \textsf{Remove-Answer}   & Strip all occurrences of the ground-truth answer from the chain. \\
$\mathcal{R}_{q}$      & \textsf{Remove-Question} & Omit the original question from the answer-generation prompt. \\
\midrule
$\mathcal{N}_{k}$ & \textsf{Noise-Injection} ($k\!\times$) & Insert the false-answer sentence at $k\times$ the frequency of the true answer. \\
\midrule
$\mathcal{D}_\text{tok}$ & \textsf{Random-Token} & Replace each token with a uniformly sampled random token. \\
$\mathcal{D}_\text{word}$ & \textsf{Random-Word}  & Replace each word with a word sampled according to the word-frequency distribution of the reasoning chain. \\
\bottomrule
\end{tabular}
\end{footnotesize}
\end{center}
\end{table*}

%% file: sections/results.tex

\section{Results and Discussion}
\label{sec:results}

We report seven empirical findings organized around four questions:
(1) which evaluation mode cleanly isolates answer extraction from independent reasoning (\Cref{sec:results-genret});
(2) whether extraction requires the reasoning chain to be presented in its original sequential order (\Cref{sec:results-order});
(3) whether informational content is uniformly distributed across the chain or concentrated in sparse symbolic structures (\Cref{sec:results-sparse});
and (4) how robust the identified extraction signal is to adversarial noise injection (\Cref{sec:results-robust}).
All accuracy values are reported under the retrieval mode (\textsf{Ret}) setting unless otherwise noted; the motivation for this choice is developed in \Cref{sec:results-genret}.

\subsection{Generation vs.\ Retrieval Mode}
\label{sec:results-genret}

We begin by motivating the choice of evaluation mode.
Our framework supports two settings: Free generation (\textsf{Gen}), in which the model freely continues its output after the reasoning chain, and retrieval mode (\textsf{Ret}), in which the model is constrained to complete a fixed answer-extraction prefix (e.g., \textit{``Thus, the answer is''}).
The distinction between these modes is consequential for interpreting any perturbation experiment.

\paragraph{Finding 1: generation mode implicitly re-reasons when the chain is absent or degraded---making retrieval mode the appropriate mode for isolating extraction.}
On the \textsf{Original} (untransformed) chain, \textsf{Gen} and \textsf{Ret} modes yield nearly identical accuracies across all three models and benchmarks (\Cref{tab:shuffle_comparison}), confirming that both modes are equally constrained by the chain's content when it is intact.
The two modes diverge dramatically, however, under the \textsf{Reasoning-Free} condition ($\mathcal{R}_r$): \texttt{OLMo-3.1-32B} reaches 28.33\% and \texttt{DeepSeek-V3.1-671B} reaches 46.33\% on \texttt{AIME 2025} in \textsf{Gen} mode \emph{without any reasoning chain}, while both collapse to 0.00\% in \textsf{Ret} mode.
This reveals a critical confound---\textbf{generation mode allows the model to substitute its own reasoning when the chain is absent or degraded}, conflating extraction with the model's native problem-solving capacity and making any measured gain uninterpretable.
We therefore adopt retrieval mode as the primary evaluation setting for all subsequent experiments, ensuring that all measured accuracy is attributable to the structure of the reasoning chain itself.

\input{tables/shuffle_comparison}
\input{tables/base_comparison}

\subsection{Order Shuffling}
\label{sec:results-order}

With the evaluation mode established, we ask whether answer extraction requires the reasoning chain to be presented in its original sequential order.
\Cref{tab:shuffle_comparison} reports accuracy under shuffling at five granularities across three models and three benchmarks.

\paragraph{Finding 2: Reasoning chains are order-shuffling at the line level, and shuffled chains provide genuine extractable signal.}
Randomly permuting the \emph{lines} of a reasoning chain (\textsf{Line-Shuffle}, $\mathcal{S}_\text{line}$)---thereby destroying all inter-line ordering while preserving each line's content---causes negligible accuracy loss.
On AIME 2025, \texttt{OLMo-3.1-32B} is unchanged at 78.00\%, \texttt{GPT-OSS-120B} drops by only 0.39 pp (91.33\% $\to$ 90.94\%), and \texttt{DeepSeek-V3.1-671B} shows a slight \emph{improvement}.
On \texttt{GPQA-Diamond}, both \texttt{GPT-OSS-120B} and \texttt{DeepSeek-V3.1-671B} record zero accuracy change under complete line-order destruction.
The sequential ordering of multi-step reasoning steps is almost entirely dispensable for answer extraction.

Crucially, this near-lossless shuffling is not because the shuffled chains are uninformative noise.
Comparing against two lower bounds---the \textsf{Reasoning-Free} baseline ($\mathcal{R}_r$, collapsing to 0.00\% for \texttt{OLMo-3.1-32B} and \texttt{DeepSeek-V3.1-671B} on \texttt{AIME 2025}) and the \textsf{Random-Token} baseline ($\mathcal{D}_\text{tok}$, reaching 0.00\%--12.54\%)---shuffled chains remain far more useful than an absent chain or purely random token sequences.
Shuffled reasoning chains preserve \textbf{genuine extractable signal}; the sequential line order, by contrast, carries almost none of the information needed for extraction.

\paragraph{Finding 3: Word-level semantic units are the minimum necessary granularity; once this minimum meaning-conveying unit is intact, ordering above it has limited impact on extraction.}
Finding 2 showed that line-level order is dispensable; Finding 3 probes how far this order-independence extends down the granularity hierarchy.
The key question is: at what level of representation does ordering begin to matter?
Cross-line word shuffling ($\mathcal{S}_\text{word}$, which scrambles all words across the entire chain) still yields substantially higher accuracy than the reasoning-free baseline: 77.67\% for \texttt{OLMo-3.1-32B}, 62.21\% for \texttt{GPT-OSS-120B}, and 89.67\% for \texttt{DeepSeek-V3.1-671B} on \texttt{AIME 2025}.
Word-level semantic content, even when globally disordered, continues to provide extractable signal---suggesting that once the minimum meaning-conveying unit is preserved, positional structure above that level exerts limited influence on extraction.

By contrast, token-level shuffling ($\mathcal{S}_\text{tok}$) is catastrophic: 1.33\% for \texttt{OLMo-3.1-32B} and 3.33\% for \texttt{GPT-OSS-120B}, approaching the reasoning-free floor.
The distinction between word shuffling and vocabulary-level randomization further isolates the effect: the \textsf{Random-Word} baseline ($\mathcal{D}_\text{word}$), which replaces each word with a sample from the original word-frequency distribution, yields 0.00\%--2.67\% on \texttt{AIME 2025}---indistinguishable from no reasoning.
Yet $\mathcal{S}_\text{word}$, which merely reorders the original words, retains 62.21\%--89.67\%.
The gap establishes that \textbf{word-level semantic identity is the minimum necessary granularity}: replacing words destroys all signal, while reordering them preserves most of it.
Any ordering structure above the word level---within-line or across-line---appears largely dispensable for answer extraction, at least within the settings we examine.

\paragraph{Finding 4: Order-independence is present in pretrained models, suggesting it originates from pretraining rather than reasoning-specific fine-tuning.}
\Cref{tab:base_comparison} compares \texttt{OLMo-3.1-32B-Base} (pretrained only) against \texttt{OLMo-3.1-32B} (instruction-tuned) under the same shuffling conditions.
Both models achieve nearly identical accuracy under \textsf{Line-Shuffle} (78.67\% vs.\ 78.00\%) and \textsf{Word-Shuffle} (78.33\% vs.\ 77.67\%) on \texttt{AIME 2025}.
Instruction tuning confers no additional advantage in handling disordered reasoning chains, and no additional penalty either.
Order-independence is therefore not a product of chain-of-thought fine-tuning but is instead rooted in the transformer architecture's pretraining regime.

Together, these results show that answer extraction is largely \textbf{order-independent above the word level}, and that this property already exists in pretrained models.

\input{tables/three_ablation_tables}

\subsection{Information Sparsity}
\label{sec:results-sparse}

Having established that sequential order does not matter, we next ask which \emph{parts} of the reasoning chain carry information.
We investigate whether the chain's content is uniformly necessary or concentrated in specific symbolic structures.

\paragraph{Finding 5: Informational content is non-uniformly distributed---alphabetic prose is redundant, the explicit answer occurrence is a strong extraction anchor, and numeric content is irreducible.}
\Cref{tab:intervention_ablation} reveals a striking asymmetry in the information structure of mathematical reasoning chains.
Masking all alphabetic characters ($\mathcal{M}_{\alpha}$, row A2) does not reduce accuracy; it \emph{increases} it by 4.7 pp (91.3\% $\to$ 96.0\%), indicating that the model relies on numeric and symbolic structures rather than on natural-language prose to locate the answer.
\Cref{tab:remove_alphabet_ablation} confirms this: \emph{removing} all alphabetic characters ($\mathcal{R}_{\alpha}$) costs only 1.3 pp (91.3\% $\to$ 90.0\%) while substantially compressing the token count.

The answer occurrence, by contrast, plays a decisive role.
Masking only the answer token(s) (\textsf{Mask-Answer}, row A3) causes an 18.0 pp drop (91.3\% $\to$ 73.3\%), showing that the explicit chain-final answer functions as a primary extraction anchor.
The model can partially recover from surrounding numeric context, but in 18.0 pp of cases the chain-final answer is the decisive cue.

Numeric content anchors the entire information hierarchy.
Masking all numeric digits ($\mathcal{M}_{\nu}$, rows A5/B5/C5/D5) collapses accuracy to 0.0\% in every condition, confirming that numeric information is both necessary and sufficient for answer extraction in mathematical reasoning.
In short: alphabetic text is redundant, the answer occurrence is a strong anchor, and numbers are irreducible.

\paragraph{Finding 6: The non-uniformly distributed informational content is sufficient on its own---perturbing its positional arrangement causes only minimal signal loss.}
Finding 5 established a non-uniform distribution of informational content: alphabetic prose is redundant, while numeric content and the explicit answer occurrence are the load-bearing components.
Finding 6 asks whether this sparse substrate remains sufficient once its positional structure is destroyed.
Row B2 of \Cref{tab:intervention_ablation} applies \textsf{Mask-Alphabet} and \textsf{Line-Shuffle} jointly: accuracy reaches 90.6\%, only $-$0.7 pp from baseline.
Stripping the prose and then permuting the remaining content causes virtually no additional loss, demonstrating that it is the \emph{content}, not its arrangement, that supports extraction.

The \textsf{Remove-Alphabet} ablation (\Cref{tab:remove_alphabet_ablation}) corroborates this under an even stronger perturbation.
Deleting all alphabetic characters and then shuffling lines ($\mathcal{S}_\text{line} \circ \mathcal{R}_{\alpha}$, row 4 of \Cref{tab:remove_alphabet_ablation}) yields 83.3\%---only $-$8.0 pp from the unperturbed baseline.
A reasoning chain stripped of all natural language and presented in arbitrary line order still extracts the correct answer in 83\% of cases.
The question prompt is similarly dispensable: removing it (rows C1, D1 of \Cref{tab:intervention_ablation}) costs at most 2.0 pp, and combining its removal with line shuffling (D1, 91.3\%) incurs zero loss.

Together, Findings 2--6 converge on a unified account: answer extraction is \textbf{order-independent above the word level} and is governed by the \textbf{non-uniform informational content}. In short, what matters is \emph{which} symbolic content is present, not \emph{where} it appears.

\subsection{Noise Robustness}
\label{sec:results-robust}

Finding 6 showed that the non-uniformly distributed informational content---numeric computations and the explicit answer occurrence---forms a self-sufficient extraction substrate whose signal survives arbitrary reordering.
A complementary question is whether this signal is also robust to adversarial contamination: when the true answer is still present but accompanied by an overwhelming number of competing false answers, can the model be misdirected away from it?
To test this, we inject the spurious sentence \textit{``Thus answer: 123.''} at $1\times$, $2\times$, and $3\times$ the true-answer occurrence count ($\mathcal{N}_{k=1}$, $\mathcal{N}_{k=2}$, $\mathcal{N}_{k=3}$) under the four representation conditions reported in \Cref{tab:noise_ablation}.

\paragraph{Finding 7: The structural signal identified in Finding 6 is intrinsically robust---even strong false-answer injection cannot override it.}
Across every representation condition that preserves numeric content, accuracy stays essentially unchanged when false answers are injected at up to $3\times$ the true-answer frequency.
The unperturbed baseline holds at 91.3\% across all noise levels ($0\times$ through $3\times$); the \textsf{Remove-Alphabet} chain remains near 90\%, and the line-shuffled chain decays only mildly (90.9\% $\to$ 86.3\% at $3\times$).
Most strikingly, even the most aggressively reduced representation from Finding 6---the line-shuffled \textsf{Remove-Alphabet} chain, retaining only a disordered numeric skeleton---stays flat at 83.3\% regardless of noise multiplier.

This robustness stands in stark contrast to the model's sensitivity to \emph{removing} the true answer (Finding 5): the model is highly sensitive to the \emph{absence} of the correct answer, yet entirely insensitive to a \emph{surplus} of competing false ones.
If extraction relied on simple statistical frequency---selecting whichever candidate appears most often---then injection at $2\times$ or $3\times$ frequency should systematically redirect the model to the noise token.
The observed immunity falsifies this account.
Instead, the non-uniform informational content of the reasoning chain---the numeric computations, relational equalities, and symbolic derivations identified in Findings 5 and 6---forms a \textbf{structurally coherent scaffold} with strong intrinsic signal: these relational patterns collectively constrain the model to the correct answer and actively suppress spurious candidates, even when false answers numerically dominate the context by frequency.

\subsection{Theoretical Conjecture} 

The pattern is consistent with transformers processing reasoning chains as approximately unordered sets of symbolic constraints, attending selectively to numeric values and relational structures while treating prose and positional ordering as low-information context. Because self-attention is architecturally near-permutation-equivariant up to positional encoding, we conjecture that \emph{relational} rather than \emph{positional} structure governs extraction---a property potentially rooted in pretraining (Finding 4) and robust to adversarial noise (Finding 7). This robustness may arise because the relational scaffold suppresses spurious candidates through implicit constraint satisfaction over the numeric subspace. However, this interpretation remains speculative and requires further experimental validation, which we leave to future work.

%% file: tables/shuffle_comparison.tex
\begin{table*}[t]
\caption{Accuracy under five shuffling granularities,
         two randomisation baselines ($\mathcal{D}_\text{tok}$, $\mathcal{D}_\text{word}$),
         and the reasoning-free condition ($\mathcal{R}_r$),
         across three models (\texttt{OLMo}, \texttt{GPT-OSS}, \texttt{DeepSeek}) and three benchmarks (\texttt{AIME 2025}, \texttt{CodeElo}, \texttt{GPQA-Diamond}).
         Two evaluation modes are compared:
         \textsf{Gen} (free generation) and
         \textsf{Ret} (retrieval with an answer-extraction prefix).
         Accuracies are single-run binomial estimates; SE\,$\leq$\,$\pm$2.9pp for \texttt{AIME} ($n=300$),
         $\pm$2.5pp for \texttt{CodeElo} ($n=408$), and $\pm$3.6pp for \texttt{GPQA-Diamond} ($n=198$).}
\label{tab:shuffle_comparison}
\begin{center}
\begin{footnotesize}

\setlength{\tabcolsep}{3pt}
\renewcommand{\arraystretch}{1.0}
\begin{tabular}{@{}rl ccc ccc ccc @{}}
\toprule
\multirow{3}{*}{~} & \multirow{3}{*}{} &
  \multicolumn{3}{c}{\texttt{AIME 2025} (Math)} &
  \multicolumn{3}{c}{\texttt{CodeElo} (Code)} &
  \multicolumn{3}{c}{\texttt{GPQA-Diamond} (Science)} \\
\cmidrule(lr){3-5} \cmidrule(lr){6-8} \cmidrule(lr){9-11}
  & & \texttt{OLMo} & \texttt{GPT-OSS} & \texttt{DeepSeek} &
    \texttt{OLMo} & \texttt{GPT-OSS} & \texttt{DeepSeek} &
    \texttt{OLMo} & \texttt{GPT-OSS} & \texttt{DeepSeek} \\
\midrule
\multirow{2}{*}{\textsf{Original Reasoning}}
  & \textsf{Gen} & 78.00 & 91.33 & 88.00 & 55.17 & 49.14 & 64.22 & 57.87 & 72.22 & 78.28 \\
  & \textsf{Ret}  & 78.00 & 91.33 & 88.67 & 55.91 & 50.25 & 63.48 & 58.08 & 71.21 & 78.79 \\

\multirow{2}{*}{\textsf{Reasoning-Free} ($\mathcal{R}_r$)}
  & \textsf{Gen} & 28.33 & 7.00 & 46.33 & 49.01 & 23.20 & 4.17 & 32.32 & 48.48 & 67.17 \\
  & \textsf{Ret}  & 0.00 & 4.00 & 0.00 & 7.64 & 21.81 & 0.00 & 35.86 & 44.44 & 46.97 \\
\midrule
\multirow{1}{*}{$\mathcal{D}_\text{tok}$}
  & \textsf{Ret}  & 0.00 & 12.54 & 0.00 & 10.84 & 13.73 & 17.89 & 28.28 & 36.87 & 49.49 \\

\multirow{1}{*}{$\mathcal{D}_\text{word}$}
  & \textsf{Ret}  & 0.00 & 3.33 & 2.67 & 14.78 & 21.32 & 20.59 & 22.22 & 38.89 & 47.47 \\
\midrule
\multirow{1}{*}{$\mathcal{S}_\text{tok}$}
  & \textsf{Ret}  & 1.33 & 3.33 & 16.33 & 12.81 & 27.45 & 20.83 & 26.77 & 44.44 & 56.06 \\

\multirow{1}{*}{$\mathcal{S}_\text{word}$ }
  & \textsf{Ret}  & 77.67 & 62.21 & 89.67 & 0.25 & 31.06 & 27.21 & 47.98 & 55.56 & 69.70 \\

\multirow{1}{*}{$\mathcal{S}_\text{line}$}
  & \textsf{Ret}  & 78.00 & 90.94 & 89.67 & 46.31 & 43.38 & 61.03 & 55.56 & 71.21 & 78.79 \\

\multirow{1}{*}{$\mathcal{S}_\text{ilw}$}
  & \textsf{Ret}  & 72.00 & 90.60 & 90.33 & 3.94 & 36.03 & 65.69 & 54.55 & 69.70 & 78.79 \\

\multirow{1}{*}{$\mathcal{S}_\text{line} \circ \mathcal{S}_\text{ilw}$}
  & \textsf{Ret}  & 72.67 & 82.07 & 89.67 & 0.49 & 29.41 & 47.30 & 56.48 & 64.65 & 76.77 \\

\bottomrule
\end{tabular}

\end{footnotesize}
\end{center}
\end{table*}

%
%

%% file: tables/base_comparison.tex
\begin{table*}[t]
\caption{Accuracy of \texttt{OLMo-Base} (pretrained only) versus \texttt{OLMo} (instruction-tuned) under identical
         shuffling conditions across three benchmarks (\textsf{Ret} setting),
         probing whether order-independence originates from pretraining or reasoning-specific fine-tuning.
         SE bounds follow Table~\ref{tab:shuffle_comparison}.}
\label{tab:base_comparison}
\begin{center}
\begin{footnotesize}
\setlength{\tabcolsep}{3pt}
\renewcommand{\arraystretch}{1.0}
\begin{tabular}{@{}rcccccc@{}}
\toprule
\multirow{3}{*}{~} &
  \multicolumn{2}{c}{\texttt{AIME 2025} (Math)} &
  \multicolumn{2}{c}{\texttt{CodeElo} (Code)} &
  \multicolumn{2}{c}{\texttt{GPQA-Diamond} (Science)} \\
\cmidrule(lr){2-3} \cmidrule(lr){4-5} \cmidrule(lr){6-7}
  & \texttt{OLMo-Base} & \texttt{OLMo} &
    \texttt{OLMo-Base} & \texttt{OLMo} &
    \texttt{OLMo-Base} & \texttt{OLMo} \\
\midrule

\textsf{Original Reasoning} & 78.33 & 78.00 & 53.20 & 55.91 & 59.60 & 58.08 \\

\textsf{Reasoning-Free} ($\mathcal{R}_r$) & 0.00 & 0.00 & 7.64 & 7.64 & 36.36 & 35.86 \\


\midrule

$\mathcal{S}_\text{word}$ & 78.33 & 77.67 & 9.85 & 0.25 & 50.51 & 47.98 \\

$\mathcal{S}_\text{line}$ & 78.67 & 78.00 & 37.93 & 46.31 & 57.07 & 55.56 \\

$\mathcal{S}_\text{ilw}$ & 78.33 & 72.00 & 34.98 & 3.94 & 59.60 & 54.55 \\

$\mathcal{S}_\text{line} \circ \mathcal{S}_\text{ilw}$ & 78.67 & 72.67 & 17.49 & 0.49 & 56.57 & 56.48 \\



\bottomrule
\end{tabular}
\end{footnotesize}
\end{center}
\end{table*}

%% file: tables/three_ablation_tables.tex

\providecommand{\onmark}{{\Large$\bullet$}}
\providecommand{\offmark}{{\Large$\circ$}}

\begin{table}[t]
\begin{minipage}[t]{0.48\linewidth}
\captionof{table}{Accuracy under combinations of five interventions---question removal ($\mathcal{R}_q$),
  alphabet masking ($\mathcal{M}_\alpha$), digit masking ($\mathcal{M}_\nu$), answer masking ($\mathcal{M}_\text{ans}$),
  and line shuffling ($\mathcal{S}_\text{line}$)---on \texttt{AIME 2025} (Math) with \texttt{GPT-OSS} (\textsf{Ret}),
  probing which components of a reasoning chain are informational.
  \onmark{} indicates the intervention is applied; \offmark{} otherwise.
  Rows are grouped by the ($\mathcal{R}_q$, $\mathcal{S}_\text{line}$) configuration:
  \textbf{A} = neither, \textbf{B} = $\mathcal{S}_\text{line}$ only,
  \textbf{C} = $\mathcal{R}_q$ only, \textbf{D} = both;
  within each group, sub-IDs 1--5 sweep the remaining three masking interventions.
  $\Delta$Acc is relative to the baseline row A1 (91.3\%).
  Light gray: $\Delta\text{Acc} < -25\%$; dark gray: $\Delta\text{Acc} < -60\%$.
  SE bounds follow Table~\ref{tab:shuffle_comparison}.}
\label{tab:intervention_ablation}
\vskip 0.15in
\centering
\begin{footnotesize}
\setlength{\tabcolsep}{4pt}
\renewcommand{\arraystretch}{1.0}
\begin{tabular}{@{}lcccccc c@{}}
\toprule
\multirow{2}{*}{ID} &
  \multicolumn{5}{c}{Transformation} &
  \multirow{2}{*}{\shortstack{Acc\\(\%)}} &
  \multirow{2}{*}{\shortstack{$\Delta$Acc\\(\%)}} \\
\cmidrule(lr){2-6}
  & $\mathcal{R}_{q}$ & $\mathcal{M}_{\alpha}$ & $\mathcal{M}_{\nu}$ & $\mathcal{M}_\text{ans}$ & $\mathcal{S}_\text{line}$ & & \\
\midrule
A1 & \offmark & \offmark & \offmark & \offmark & \offmark & 91.3 & --- \\
A2 & \offmark & \onmark  & \offmark & \offmark & \offmark & 96.0 & +4.7 \\
A3 & \offmark & \offmark & \offmark & \onmark  & \offmark & 73.3 & -18.0 \\
A4 & \offmark & \onmark  & \offmark & \onmark  & \offmark & 68.1 & -23.2 \\
\rowcolor{gray!60} A5 & \offmark & \offmark & \onmark  & \onmark  & \offmark & 0.0 & -91.3 \\
B1 & \offmark & \offmark & \offmark & \offmark & \onmark  & 90.9 & -0.4 \\
B2 & \offmark & \onmark  & \offmark & \offmark & \onmark  & 90.6 & -0.7 \\
B3 & \offmark & \offmark & \offmark & \onmark  & \onmark  & 71.3 & -20.0 \\
\rowcolor{gray!20} B4 & \offmark & \onmark  & \offmark & \onmark  & \onmark  & 51.8 & -39.5 \\
\rowcolor{gray!60} B5 & \offmark & \offmark  & \onmark & \onmark & \onmark  & 0.0 & -91.3 \\
C1 & \onmark  & \offmark & \offmark & \offmark & \offmark & 89.3 & -2.0 \\
C2 & \onmark  & \onmark  & \offmark & \offmark & \offmark & 92.8 & +1.5 \\
\rowcolor{gray!20} C3 & \onmark & \offmark & \offmark & \onmark  & \offmark & 63.7 & -27.6 \\
\rowcolor{gray!60} C4 & \onmark  & \onmark  & \offmark & \onmark  & \offmark & 18.8 & -72.5 \\
\rowcolor{gray!60} C5 & \onmark & \offmark  & \onmark & \onmark & \offmark  & 0.0 & -91.3 \\
D1 & \onmark  & \offmark & \offmark & \offmark & \onmark  & 91.3 & 0.0 \\
\rowcolor{gray!20} D2 & \onmark  & \onmark  & \offmark & \offmark & \onmark  & 54.0 & -37.3 \\
\rowcolor{gray!20} D3 & \onmark & \offmark & \offmark & \onmark  & \onmark  & 65.67 & -25.6 \\
\rowcolor{gray!60} D4 & \onmark & \onmark  & \offmark & \onmark  & \onmark  & 3.62 & -87.7 \\
\rowcolor{gray!60} D5 & \onmark & \offmark  & \onmark & \onmark & \onmark  & 0.0 & -91.3 \\
\bottomrule
\end{tabular}
\end{footnotesize}
\end{minipage}
\hfill
\begin{minipage}[t]{0.48\linewidth}

  \captionof{table}{Accuracy when all alphabetic characters are removed ($\mathcal{R}_{\alpha}$),
    combined with answer removal ($\mathcal{R}_\text{ans}$), line shuffling ($\mathcal{S}_\text{line}$),
    or word shuffling ($\mathcal{S}_\text{word}$), on \texttt{AIME 2025} (Math) with \texttt{GPT-OSS} (\textsf{Ret}).
    $\Delta$Acc is relative to the baseline (91.3\%).
    Light gray: $\Delta\text{Acc} < -25\%$; dark gray: $\Delta\text{Acc} < -60\%$.
    Example chains under each transformation appear in
    Table~\ref{tab:remove_alphabet_examples}.
    SE bounds follow Table~\ref{tab:shuffle_comparison}.}
  \label{tab:remove_alphabet_ablation}
  \vskip 0.1in
  \centering
  \begin{footnotesize}
  \setlength{\tabcolsep}{5pt}
  \renewcommand{\arraystretch}{1.0}
  \begin{tabular}{@{}lcc@{}}
  \toprule
  Transformation & Acc (\%) & $\Delta$Acc (\%) \\
  \midrule
  \textsf{Original}                                                 & 91.3 & ---   \\
  $\mathcal{R}_{\alpha}$                                            & 90.0 & -1.3  \\
  \rowcolor{gray!20}$\mathcal{R}_{\alpha} \circ \mathcal{R}_\text{ans}$              & 58.7 & -32.6 \\
  $\mathcal{S}_\text{line} \circ \mathcal{R}_{\alpha}$              & 83.3 & -8.0  \\
  \rowcolor{gray!20}$\mathcal{S}_\text{line} \circ \mathcal{R}_{\alpha} \circ \mathcal{R}_\text{ans}$ & 46.7 & -44.6 \\
  \rowcolor{gray!20}$\mathcal{S}_\text{word} \circ \mathcal{R}_{\alpha}$             & 37.3 & -54.0 \\
  \bottomrule
  \end{tabular}
  \end{footnotesize}

  \vspace{2em}

  \captionof{table}{Accuracy under false-answer noise injection ($\mathcal{N}_k$) on \texttt{AIME 2025}
    (Math) with \texttt{GPT-OSS} (\textsf{Ret}).
    Rows cross six representation conditions: three shuffle states (none,
    $\mathcal{S}_\text{line}$, $\mathcal{S}_\text{word}$) crossed with alphabet removal
    ($\mathcal{R}_{\alpha}$) on/off;
    columns vary the noise multiplier from $0\times$ to $3\times$ the true-answer occurrence count.
    SE bounds follow Table~\ref{tab:shuffle_comparison}.}
  \label{tab:noise_ablation}
  \centering
  \begin{footnotesize}
  \setlength{\tabcolsep}{4pt}
  \renewcommand{\arraystretch}{1.0}
  \begin{tabular}{@{}cc cccc@{}}
  \toprule
  \multirow{2}{*}{Shuffle} & \multirow{2}{*}{\shortstack{Remove-\\Alphabet}}
    & \multicolumn{4}{c}{Acc (\%) by Noise Multiplier} \\
  \cmidrule(l){3-6}
   &  & $-$ & $\mathcal{N}_{k=1}$ & $\mathcal{N}_{k=2}$ & $\mathcal{N}_{k=3}$ \\
  \midrule
  $-$                       & $-$                  & 91.3 & 91.6 & 91.6 & 91.3 \\
  $-$                       & $\mathcal{R}_{\alpha}$ & 90.0 & 90.7 & 90.7 & 90.0 \\
  $\mathcal{S}_\text{line}$ & $-$                  & 90.9 & 90.0 & 87.3 & 86.3 \\
  $\mathcal{S}_\text{line}$ & $\mathcal{R}_{\alpha}$ & 83.3 & 83.3 & 83.3 & 83.3 \\
  $\mathcal{S}_\text{word}$ & $-$                  & 62.2 & 61.5 & 64.2 & 64.9 \\
  $\mathcal{S}_\text{word}$ & $\mathcal{R}_{\alpha}$ & 37.3 & 35.3 & 27.0 & 30.3 \\
  \bottomrule
  \end{tabular}
  \end{footnotesize}

\end{minipage}
\end{table}

%% file: sections/conclusion.tex

\section{Conclusion}
\label{sec:conclusion}

Modern reasoning language models generate dense, sequential chain-of-thought traces under two implicit but untested assumptions---that every token contributes (\emph{density}) and that steps must be consumed in order (\emph{order}).
We asked what information the answer-\emph{extraction} stage actually relies on, and answered empirically through a systematic intervention pipeline---removal, masking, shuffling, and noise injection. 

Our seven findings overturn both assumptions along three converging dimensions.
\emph{Order}: line-level shuffling reduces accuracy by less than 0.5 pp, word-level shuffling still retains 62\%--89\%, and pretrained-only and instruction-tuned variants tolerate shuffling near-identically---indicating that order-independence originates from pretraining rather than reasoning-specific fine-tuning.
\emph{Dense}: masking alphabetic prose \emph{improves} accuracy by 4.7 pp, the answer is a strong anchor (an 18.0 pp drop when masked alone), and masking digits collapses accuracy to 0\%.
\emph{Robustness}: a chain stripped of all natural language and shuffled into arbitrary line order still yields 83\%, and false-answer injection at $3\times$ frequency leaves accuracy unchanged, falsifying a frequency-based extraction account.
Together, these results establish that answer extraction operates on a \textbf{sparse, order-shuffling, and structurally robust informational substrate}, opening concrete paths toward parallelized and token-efficient reasoning.


\textbf{Limitations.} This work probes only answer \emph{extraction}; whether generation can be made commensurately sparse or parallel is an open engineering question we motivate but do not resolve. Sparsity findings rest primarily on \texttt{AIME 2025}, whose numeric-essential structure is inherent to mathematical notation and may differ for code or scientific reasoning. Order-independence holds strictly above the word level, and word-shuffle tolerance varies non-trivially across models. Our theoretical conjecture is consistent with the data; however, it lacks mechanistic verification and requires further experimental validation.

\textbf{Impact Statement.} The primary benefit is more token-efficient and parallelizable inference, lowering deployment cost. We foresee no direct harmful applications of this research.

%% file: tables/remove_alphabet_examples.tex
\begingroup

\refstepcounter{table}
\label{tab:remove_alphabet_examples}
\noindent\textbf{Table~\thetable:} Example reasoning chains for the baseline and the five
transformations in Table~\ref{tab:remove_alphabet_ablation}, applied to the question:
\textit{Find the sum of all integer bases $b>9$ for which $17_{b}$ is a divisor of $97_{b}$.}
Correct answer highlighted in \textcolor{red}{red}.
\par\addvspace{0.6em}
\hrule
\par\addvspace{0.3em}

\noindent Baseline\par\addvspace{0.2em}
{\scriptsize
We are asked: Find the sum of all integer bases b $>$ 9 for which 17\_b divides 97\_b.\par\smallskip
Interpretation: In base b, the numeral "17" means the integer 1*b + 7 = b + 7. Similarly, "97" in base b means 9*b + 7 = 9b + 7. So the condition is that (b + 7) divides (9b + 7). For b $>$ 9 (since digits must be less than base, digits used are 1,7,9,7. So base must be $>$ 9 to allow digit "9").\par\smallskip
We need integer bases b $>$ 9 such that (b + 7) $|$ (9b + 7). Find all b $>$ 9 integer solutions.\par\smallskip
So we solve (9b + 7) mod (b + 7) = 0.\par\smallskip
Compute remainder: 9b + 7 = 9(b + 7) - 9*7 + 7 = 9(b + 7) - 63 + 7 = 9(b + 7) - 56.\par\smallskip
Thus remainder when dividing by b + 7 is -56 (or equivalently b+7 - 56 mod). So we want b+7 divides 56. Since remainder is -56, we need b+7 divides 56. But note that remainder of division is 9b+7 - 9(b+7) = -56. So 9b+7 = 9(b+7) - 56. If b+7 divides 9(b+7) obviously, then b+7 will also divide the remainder -56. So condition: b+7 divides 56. But also b+7 $>$ 0.\par\smallskip
Thus b+7 is a positive divisor of 56.\par\smallskip
Since b $>$ 9 =$>$ b+7 $>$ 16. So positive divisors of 56 greater than 16.\par\smallskip
Divisors of 56: 1,2,4,7,8,14,28,56.\par\smallskip
Those $>$ 16: 28,56.\par\smallskip
Thus b+7 can be 28 or 56.\par\smallskip
So b = 21 or 49. Both $>$9, integer.\par\smallskip
Check base digits constraints: need digits 9,7 allowed. Base must be $>$9, which both satisfy.\par\smallskip
Thus the required sum is 21+49 = \textcolor{red}{70}.\par\smallskip
Wait check also b+7 could be negative? No base is positive integer $>$9.\par\smallskip
Thus answer \textcolor{red}{70}.\par\smallskip
But perhaps also b+7 dividing -56 (which is same). But we also need to ensure that b+7 indeed divides 9b+7 (i.e., remainder zero). Our condition implies b+7 divides 56. Since -56 remainder is multiple of divisor; but careful: remainder must be 0, i.e., -56 should be multiple of b+7. So b+7 is a divisor of 56. Should -56 be exactly divisible: b+7 $|$ 56. So that's the condition.\par\smallskip
Hence solutions as above.\par\smallskip
Double-check those bases indeed work:\par\smallskip
b=21: 17\_21 = 21+7=28; 97\_21 = 9*21+7 = 189+7 = 196. 196/28 = 7 -$>$ integer.\par\smallskip
b=49: 17\_49 = 49+7 = 56; 97\_49 = 9*49+7 = 441+7=448. 448/56 =8 -$>$ integer.\par\smallskip
Thus sum = \textcolor{red}{70}.\par\smallskip
Thus answer is \textcolor{red}{70}.\par\smallskip
But could there be any other possibilities? Let's check for b+7 dividing 56 also could be 1? But b+7 must be $>$ 16 as we said. So only 28,56. Verified.\par\smallskip
Thus answer \textcolor{red}{70}.\par\smallskip
However, is there any catch that digits must be less than base, but base 21 and 49 satisfy digits max 9 $<=$ base-1.\par\smallskip
Thus answer stands.\par\smallskip
Thus solution: \textcolor{red}{70}.\par\smallskip
In final answer format: \textcolor{red}{70}.\par\smallskip
Will provide reasoning.\par\smallskip
Proceed to final.
\par}

\par\addvspace{0.8em}\hrule\par\addvspace{0.3em}

\noindent $\mathcal{R}_{\alpha}$ \par\addvspace{0.2em}
{\scriptsize
: $>$ 9 17\_ 97\_ . : , "17" 1* + 7 = + 7. , "97" 9* + 7 = 9 + 7. ( + 7) (9 + 7). $>$ 9 ( , 1,7,9,7. $>$ 9 "9"). $>$ 9 ( + 7) $|$ (9 + 7). $>$ 9 . (9 + 7) ( + 7) = 0. : 9 + 7 = 9( + 7) - 9*7 + 7 = 9( + 7) - 63 + 7 = 9( + 7) - 56. + 7 -56 ( +7 - 56 ). +7 56. -56, +7 56. 9 +7 - 9( +7) = -56. 9 +7 = 9( +7) - 56. +7 9( +7) , +7 -56. : +7 56. +7 $>$ 0. +7 56. $>$ 9 =$>$ +7 $>$ 16. 56 16. 56: 1,2,4,7,8,14,28,56. $>$ 16: 28,56. +7 28 56. = 21 49. $>$9, . : 9,7 . $>$9, . 21+49 = \textcolor{red}{70}. +7 ? $>$9. \textcolor{red}{70}. +7 -56 ( ). +7 9 +7 ( . ., ). +7 56. -56 ; : 0, . ., -56 +7. +7 56. -56 : +7 $|$ 56. ' . . - : =21: 17\_21 = 21+7=28; 97\_21 = 9*21+7 = 189+7 = 196. 196/28 = 7 -$>$ . =49: 17\_49 = 49+7 = 56; 97\_49 = 9*49+7 = 441+7=448. 448/56 =8 -$>$ . = \textcolor{red}{70}. \textcolor{red}{70}. ? ' +7 56 1? +7 $>$ 16 . 28,56. . \textcolor{red}{70}. , , 21 49 9 $<=$ -1. . : \textcolor{red}{70}. : \textcolor{red}{70}. . .
\par}

\par\addvspace{0.8em}\hrule\par\addvspace{0.3em}

\noindent $\mathcal{R}_{\alpha} \circ \mathcal{R}_\text{ans}$ \par\addvspace{0.2em}
{\scriptsize
: $>$ 9 17\_ 97\_ . : , "17" 1* + 7 = + 7. , "97" 9* + 7 = 9 + 7. ( + 7) (9 + 7). $>$ 9 ( , 1,7,9,7. $>$ 9 "9"). $>$ 9 ( + 7) $|$ (9 + 7). $>$ 9 . (9 + 7) ( + 7) = 0. : 9 + 7 = 9( + 7) - 9*7 + 7 = 9( + 7) - 63 + 7 = 9( + 7) - 56. + 7 -56 ( +7 - 56 ). +7 56. -56, +7 56. 9 +7 - 9( +7) = -56. 9 +7 = 9( +7) - 56. +7 9( +7) , +7 -56. : +7 56. +7 $>$ 0. +7 56. $>$ 9 =$>$ +7 $>$ 16. 56 16. 56: 1,2,4,7,8,14,28,56. $>$ 16: 28,56. +7 28 56. = 21 49. $>$9, . : 9,7 . $>$9, . 21+49 = . +7 ? $>$9. . +7 -56 ( ). +7 9 +7 ( . ., ). +7 56. -56 ; : 0, . ., -56 +7. +7 56. -56 : +7 $|$ 56. ' . . - : =21: 17\_21 = 21+7=28; 97\_21 = 9*21+7 = 189+7 = 196. 196/28 = 7 -$>$ . =49: 17\_49 = 49+7 = 56; 97\_49 = 9*49+7 = 441+7=448. 448/56 =8 -$>$ . = . . ? ' +7 56 1? +7 $>$ 16 . 28,56. . . , , 21 49 9 $<=$ -1. . : . : . . .
\par}

\par\addvspace{0.8em}\hrule\par\addvspace{0.3em}

\noindent $\mathcal{S}_\text{line} \circ \mathcal{R}_{\alpha}$ \par\addvspace{0.2em}
{\scriptsize
=21: 17\_21 = 21+7=28; 97\_21 = 9*21+7 = 189+7 = 196. 196/28 = 7 -$>$ . +7 ? $>$9. +7 28 56. \textcolor{red}{70}. : \textcolor{red}{70}. +7 56. + 7 -56 ( +7 - 56 ). +7 56. -56, +7 56. 9 +7 - 9( +7) = -56. 9 +7 = 9( +7) - 56. +7 9( +7) , +7 -56. : +7 56. +7 $>$ 0. : 9,7 . $>$9, . = 21 49. $>$9, . \textcolor{red}{70}. $>$ 16: 28,56. . . \textcolor{red}{70}. +7 -56 ( ). +7 9 +7 ( . ., ). +7 56. -56 ; : 0, . ., -56 +7. +7 56. -56 : +7 $|$ 56. ' . : \textcolor{red}{70}. : , "17" 1* + 7 = + 7. , "97" 9* + 7 = 9 + 7. ( + 7) (9 + 7). $>$ 9 ( , 1,7,9,7. $>$ 9 "9"). 21+49 = \textcolor{red}{70}. - : $>$ 9 ( + 7) $|$ (9 + 7). $>$ 9 . . = \textcolor{red}{70}. . : 9 + 7 = 9( + 7) - 9*7 + 7 = 9( + 7) - 63 + 7 = 9( + 7) - 56. , , 21 49 9 $<=$ -1. $>$ 9 =$>$ +7 $>$ 16. 56 16. 56: 1,2,4,7,8,14,28,56. ? ' +7 56 1? +7 $>$ 16 . 28,56. . : $>$ 9 17\_ 97\_ . (9 + 7) ( + 7) = 0. =49: 17\_49 = 49+7 = 56; 97\_49 = 9*49+7 = 441+7=448. 448/56 =8 -$>$ .
\par}

\par\addvspace{0.8em}\hrule\par\addvspace{0.3em}

\noindent $\mathcal{S}_\text{line} \circ \mathcal{R}_{\alpha} \circ \mathcal{R}_\text{ans}$ \par\addvspace{0.2em}
{\scriptsize
=21: 17\_21 = 21+7=28; 97\_21 = 9*21+7 = 189+7 = 196. 196/28 = 7 -$>$ . +7 ? $>$9. +7 28 56. . : . +7 56. + 7 -56 ( +7 - 56 ). +7 56. -56, +7 56. 9 +7 - 9( +7) = -56. 9 +7 = 9( +7) - 56. +7 9( +7) , +7 -56. : +7 56. +7 $>$ 0. : 9,7 . $>$9, . = 21 49. $>$9, . . $>$ 16: 28,56. . . . +7 -56 ( ). +7 9 +7 ( . ., ). +7 56. -56 ; : 0, . ., -56 +7. +7 56. -56 : +7 $|$ 56. ' . : . : , "17" 1* + 7 = + 7. , "97" 9* + 7 = 9 + 7. ( + 7) (9 + 7). $>$ 9 ( , 1,7,9,7. $>$ 9 "9"). 21+49 = . - : $>$ 9 ( + 7) $|$ (9 + 7). $>$ 9 . . = . . : 9 + 7 = 9( + 7) - 9*7 + 7 = 9( + 7) - 63 + 7 = 9( + 7) - 56. , , 21 49 9 $<=$ -1. $>$ 9 =$>$ +7 $>$ 16. 56 16. 56: 1,2,4,7,8,14,28,56. ? ' +7 56 1? +7 $>$ 16 . 28,56. . : $>$ 9 17\_ 97\_ . (9 + 7) ( + 7) = 0. =49: 17\_49 = 49+7 = 56; 97\_49 = 9*49+7 = 441+7=448. 448/56 =8 -$>$ .
\par}

\par\addvspace{0.8em}\hrule\par\addvspace{0.3em}

\noindent $\mathcal{S}_\text{word} \circ \mathcal{R}_{\alpha}$ \par\addvspace{0.2em}
{\scriptsize
+ - 9 ( = 49+7 ). ? 448/56 9 7). 1,2,4,7,8,14,28,56. +7 -56 -56. $|$ 9* 7). , $>$ 56: 196. 0. 7 1,7,9,7. 17\_21 = ( 56 =21: 9*21+7 = + 7) - +7 ). -56 : = $>$ 16. +7 7) +7 ( $>$ ( . ., + 56. . ., , 17\_49 1? = 189+7 . - = : + + : : + 0, 16: 56. 56. 16 9( +7) = = -56 . - 21 +7 21+7=28; 7 97\_ . $>$ 9*7 +7 7 ; 9( +7) 7) 56. +7 56. = 56. + $|$ ' $>$9, \textcolor{red}{70}. + = \textcolor{red}{70}. 49. \textcolor{red}{70}. +7 7) = \textcolor{red}{70}. \textcolor{red}{70}. $<=$ 97\_49 =8 9 $>$ . 56. 97\_21 $>$ , $>$ 56. +7. = ). 7 . : 7. $>$9, 7 : -$>$ ( =$>$ 49 +7 56. , "9"). . ( $>$ 7 \textcolor{red}{70}. + + 9 ( 28,56. 9 +7 196/28 + 9 9*49+7 9 28,56. =49: + 7) : 56 = 56. +7 \textcolor{red}{70}. . "17" + +7 = , 9( 21 9( : 16. +7 28 ' +7 + - 56 -1. 7) 1* +7 . 63 441+7=448. 56; 9,7 = -56 "97" -56 7) : . -56. - 9 +7 +7 ? = + , (9 9 +7 9( +7) 9 . . . + +7 - 9 -56, = $>$ -$>$ = 9( $>$ + +7 : 21+49 0. (9 7. 17\_ +7 $>$9. = . . = 7 9 (9
\par}

\par\addvspace{0.4em}\hrule

\endgroup